\documentclass[12pt]{article}
\usepackage{graphicx}
\usepackage{hyperref}
 \usepackage{setspace}
 \usepackage{float}
 \usepackage{cite}

\usepackage{scicite}

\usepackage{times}

\newenvironment{sciabstract}{%
\hspace{-3cm}
\begin{quote} \bf}
{\end{quote}}

\newcounter{lastnote}

\title{Art and the science of generative AI: A deeper dive}

\author
{Ziv Epstein$^{1\ast}$, Aaron Hertzmann$^{2}$, Laura Herman$^{3,4}$, Robert Mahari$^{1,5}$, \\ Morgan R. Frank $^{6}$, Matthew Groh$^{1}$, Hope Schroeder$^{1}$, Amy Smith$^{7}$, \\ Memo Akten$^{8}$, Jessica Fjeld$^{5}$,
Hany Farid$^{9}$, Neil Leach$^{10}$, \\ Alex ``Sandy'' Pentland$^{1}$, Olga Russakovsky$^{11}$ \\
\normalsize{$^{1}$MIT Media Lab}\\
\normalsize{$^{2}$Adobe Research}\\
\normalsize{$^{3}$University of Oxford}\\
\normalsize{$^{4}$Adobe, Inc.}\\
\normalsize{$^{5}$Harvard Law School}\\
\normalsize{$^{6}$University of Pittsburgh}\\
\normalsize{$^{7}$Queen Mary University London}\\
\normalsize{$^{8}$University of California, San Diego}\\
\normalsize{$^{9}$University of California, Berkeley}\\
\normalsize{$^{10}$Florida International University}\\
\normalsize{$^{11}$Princeton University}\\
\\
\normalsize{$^\ast$To whom correspondence should be addressed; E-mail:  zive@mit.edu.
}
}

\date{\vspace{-1.5cm}}

\begin{document} 


\baselineskip12pt


\maketitle 
\begin{singlespace}

\begin{sciabstract}
A new class of tools, colloquially called \textit{generative AI}, can produce high-quality artistic media for visual arts, concept art, music, fiction, literature, video, and animation. The generative capabilities of these tools are likely to fundamentally alter the creative processes by which creators formulate ideas and put them into production. As creativity is reimagined, so too may be many sectors of society.  Understanding the impact of generative AI—and making policy decisions around it—requires new interdisciplinary scientific inquiry into culture, economics, law, algorithms, and the interaction of technology and creativity. We argue that generative AI is not the harbinger of art’s demise, but rather is a new medium with its own distinct affordances.  In this vein, we consider the impacts of this new medium on creators across four themes: aesthetics and culture, legal questions of ownership and credit, the future of creative work, and impacts on the contemporary media ecosystem. Across these themes, we highlight key research questions and directions to inform policy and beneficial uses of the technology.
\end{sciabstract}


Note: This white paper is an expanded version of Epstein et al 2023 published in \textit{Science} Perspectives on July 16, 2023 which you can find at the following DOI: 10.1126/science.adh4451.

\section{Introduction}
Generative AI systems increasingly have the capability to produce high-quality artistic media for visual arts, concept art, music, fiction, literature, and video/animation. For example, diffusion models can synthesize high-quality images \cite{rombach2022high} and large language models can produce sensible-sounding and impressive prose and verse in a wide range of contexts \cite{vaswani2017attention}. The generative capabilities of these tools are likely to fundamentally alter the creative processes by which creators formulate ideas and put them into production. As creativity is reimagined, so too may be many sectors of society. Understanding the impact of generative AI—and making policy decisions around it—requires new interdisciplinary scientific inquiry into culture, economics, law, algorithms, and the interaction of technology and creativity.

Generative AI tools, at first glance, seem to fully automate artistic production—an impression that mirrors past instances when traditionalists viewed new technologies as threatening “art itself.” In fact, these moments of technological change did not indicate the “end of art,” but had much more complex effects, recasting the roles and practices of creators and shifting the aesthetics of contemporary media \cite{hertzmann2018can}. For example, some 19th-century artists saw the advent of photography as a threat to painting. Instead of replacing painting, however, photography eventually liberated it from realism, giving rise to Impressionism and the Modern Art movement.  On the other hand, portrait photography did largely replace portrait painting, leading to a short-term loss of jobs among portraiturists and postcard painters \cite{scharf1974art}. 
Many other historical analogies illustrate similar trends, with a new artistic technology challenging traditional creative practices and jobs while in time creating new roles for and genres of art. The digitization of music production (e.g., digital sampling and sound synthesis) was decried as ``the end of music." Instead, it altered the ways we produce and listen to music and helped spawn new genres, like Hip Hop and Drum'n'bass. This follows trends in computer animation (where traditional animators thought that computers would replace animators entirely, but instead computer animation flourished as a medium and  jobs for computer animators increased \cite{paik, sito2013moving}) and digital photography (which in its time challenged photographic principles and assumptions, but now it is commonplace and widely used \cite{cotton2020photograph, brit}).

Like these historical analogs, generative AI is not necessarily the harbinger of art’s demise, but rather is a new medium with its own distinct affordances.  As a suite of tools used by human creators, generative AI is positioned to upend many sectors of the creative industry and beyond—threatening existing jobs and labor models in the short term, while ultimately enabling new models of creative labor and reconfiguring the media ecosystem. These immediate impacts require serious consideration and discussion across academia, industry and civil society. 

Unlike past disruptions, however, generative AI relies on training data made by people: the models “learn” to generate art by extracting statistical patterns from existing artistic media. This reliance on training data raises new issues---such as where the data is sourced, how it influences the resulting outputs, and how to determine authorship. By leveraging existing work to automate aspects of the creative process, generative AI challenges conventional definitions of authorship, ownership, creative inspiration, sampling, and remixing and thus complicates existing conceptions of media production.  It is therefore important to consider generative AI’s impacts on aesthetics and culture, legal questions of ownership and credit, the future of the creative work, and impacts on the contemporary media ecosystem. Across these themes, there are key research questions to inform policy and beneficial uses of this technology that we outline in this white paper.
\section{Perceptions and Conceptualization of Generative AI}
To properly study these themes, it is first necessary to understand how the language used to describe AI affects perceptions of the technology, as the very term “artificial intelligence” itself is fraught with traps and fallacies \cite{mitchell2021ai}. Further, many different algorithms---ranging from linear regressions and symbolic logic to large language models (LLMs)---all fall under the broad umbrella of ``artificial intelligence'' despite different levels of computational complexity and variable problem domains.

The terms and design of AI might misleadingly imply that these systems exhibit human-like intent, agency, or even self-awareness. The natural language-based interfaces that now accompany generative AI models (e.g., chat interfaces) as well as the models' usage of the ``I'' pronoun may give users a sense of human-like interaction. Furthermore, terms like “hallucination” that refers to patently false statements by large language models doubles down on this anthropomorphic frame \cite{lipton2018troubling}. 

Such design decisions can cause people to anthropomorphize AI systems, attributing human-like characteristics to them \cite{reeves1996media}. Anthropomorphizing AI can pose challenges to the ethical usage of this technology \cite{watson2019rhetoric}. In particular, perceptions of human-like agency can undermine credit to the creators whose labor underlies the system’s outputs \cite{epstein2020gets} and deflect responsibility from developers and decision-makers when these systems cause harm \cite{elish2019moral}. We, therefore, discuss generative AI as a tool to support human creators \cite{cohen2000imagination}, rather than an agent capable of harboring its own intent or authorship. In this view, there is little room for autonomous machines being ``artists'' or ``creative'' in their own right. 
 
To anchor this view, we draw on the concept of meaningful human control (MHC), a concept originally adapted from autonomous weapons literature that refers to a human operator's control over and responsibility for a computational system \cite{santoni2018meaningful}. Akten \cite{akten2021deep} adapts the MHC framework into the context of AI art to ground discussions of intent, predictability, and accountability. In order to be considered meaningful human control, a generative system should be capable of incorporating a human author's intent into its output. If a user starts with no specific goal, the system should allow for open-ended, curiosity-driven exploration. As the user's goal becomes clearer through interaction, the system should be able to both guide and deliver this intent. Such systems should have a degree of predictability, allowing users to gradually understand the system to the extent that they can learn to anticipate the results of their actions. Given these conditions, we can consider the human user as accountable for the outputs of the generative system. In other words, MHC is achieved if human creators can creatively express themselves through the generative system, leading to an outcome that aligns with their intentions and carries their personal, expressive signature. Future work is needed to investigate in what ways generative systems and interfaces can be developed that allow more meaningful human control by adding input streams that provide users fine-grained causal manipulation over outputs.  

Generative AI systems are diffuse, sociotechnical systems \cite{seaver2017algorithms} and therefore more work is needed to understand how people reason about the complex interplay between human actors and computational processes. For example, how do perceptions of the generative process (e.g., the relative salience or invisibility of various stakeholders or the involvement of AI disclosure) affect attitudes towards artifacts produced by those systems \cite{raj2023art}? And how do these perceptions affect attitudes towards various stakeholders involved in the generative AI systems in the first place \cite{epstein2020gets}? These insights can help us design systems that properly disclose the generative process and avoid misleading interpretations. 

So far, this discussion of generative AI has centered on the self-contained case in which a user queries a generative AI model directly (e.g. via prompting) to create a static artifact via inference. However, other regimes within AI art involve the development of systems that go beyond this fixed user-prompting paradigm. Many of these systems use feedback from a large number of users  to guide the creation of content \cite{epstein2020interpolating, gordon2022co, klingemann2021botto,draves, secretan2011picbreeder, kogan2019artist}, and thus fall into the lineage of collective or crowd art (e.g. the r/Place experiment \cite{rappaz2018latent} or Agnieszka Kurant's \textit{The End of Signature}\cite{shanmugam2022intersection}. Given that unforeseen outputs from AI systems can fuel perceptions and fear of AI agency, the well-established category of human-created generative art systems with intentionally unexpected outputs provide an important reminder: humans create these systems and are responsible for their outputs.
\section{Shifts in Culture \& Aesthetics}
Every artistic medium is inextricably linked to the cultural zeitgeist within which it operates, and contemporary AI-generated art itself is a reflection of contemporary issues surrounding the attention economy and corporate control. Generative AI has distinct cultural context and material affordances that make it a unique new artform (see \cite{cetinic2022understanding} for a review). First, the democratization of digital creative tools coupled with the popularity of posting images on social media has created a media ecosystem where anyone can create (and share) AI-generated content with very low barriers to entry. Second, this content is viewed on algorithmically-mediated social media platforms, where attention is scarce and explicitly monetized \cite{poell2021platforms,hwang2020subprime}.  Finally, the vast computational infrastructure necessary to produce AI-generated content (e.g. large numbers of GPUs) is developed and maintained by a few large companies, who can therefore control the functionality of and access to the technology.

These unique affordances in turn give rise to a medium with its own aesthetics that may have a long-term effect on art and culture \cite{manovich2018ai}. Primarily, we note how this medium will recast the practices and roles of creators. In traditional artforms characterized by direct manipulation \cite{shneiderman1981direct} of a material (e.g., painting, tattoo, or sculpture), the creator has a direct hand in creating the final output, and therefore it is relatively straightforward to identify the creator’s intentions and style in the output. Indeed, previous research has shown the relative importance of ``intention guessing" in the artistic viewing experience \cite{bloom1996intention,snapper2015your}, as well as the increased creative value afforded to an artwork if elements of the human process (e.g., brushstrokes) are visible \cite{herman2022eye}. 

However, generative techniques have strong aesthetics themselves \cite{manovich2021artificial}; for instance, it has become apparent that certain generative tools are built to be as ``realistic" as possible, resulting in a hyperrealistic aesthetic style. As these aesthetics propagate through visual culture, it can be difficult for a casual viewer to identify the creator's intention and individuality within the outputs. Indeed, some creators have spoken about the challenges of getting generative AI models to produce images in new, different, or unique aesthetic styles \cite{manovich2021artificial, akten2018machine}. This unique position of the creator relative to the tool calls into question the particular role of the creator in exerting their artistic intention on AI-generated artifacts. While there is a long history of generative and computer art, these art forms usually involve software built by the artist with distinctive aesthetics.

Therefore, AI-based artists using generative AI systems must find ways to exert their artistic intention and rigor into other stages of the creation process, such as how they select training data, craft prompts \cite{liu2022design, mccormack2023writing}, or use AI-generated artifacts for downstream creative applications \cite{smith2023trash}. Future work should explore what constitutes meaningful human control in the context of generative AI. How does it relate to intent, predictability, accountability and expression? What existing interactions with generative AI are sites for artistic agency and meaningful human control? How can additional sites of artistic agency and meaningful human control be introduced into generative AI systems, such as through increased explainability, transparency and responsiveness? These explorations can be organized into distinct layers like the user-facing interface layer, (i.e., user-experience design), but also a deeper layer of incorporating particular desired controls into the models themselves. 

As generative AI tools become more widespread, and knowledge of these tools becomes commonplace (as consumer photography did a century ago), an open question remains regarding how the aesthetics of their outputs will affect the range of artistic outputs. On one hand, generative AI could increase the overall diversity of artistic outputs by expanding the set of creators who engage with artistic practice.

But on the other hand, the aesthetic and cultural norms and biases embedded in the training data of generative-AI models affect their outputs. It is well documented that biases in the training data of an algorithmic system can create outputs that reflect or even amplify those biases~\cite{zhao2017men,barocas-hardt-narayanan}. The data used to train generative-AI models primarily comes from the web; web image search results have been shown to amplify existing racial and gender inequalities~\cite{noble2018algorithms,kay2015unequal}, and further be geographically concentrated rather than representative of all cultures~\cite{shankar2017no}. Without documentation of what data is used in the training of generative-AI models, it is more difficult to identify, quantify and mitigate the biases the models have~\cite{datasheets,bender2021dangers}, although efforts have been made to overcome this issue~\cite{grover2019bias,luccioni2023stable}. Going beyond the data, algorithmic decisions, such as which types of outputs to reward when training the model, implicitly reflect the values of the generative AI creators~\cite{ganrace}. For example, models may learn to produce outputs that more closely mimic the ``common'' rather than the ``rare'' or ``unique'' inputs, or focus on representing just a subset of the data~\cite{moderegularized}. Thus, it is possible that the generative AI models will in fact entrench bias in cultural production and decrease aesthetic diversity.

AI-generated content may also feed future generative models, creating a self-referential aesthetic flywheel that could perpetuate AI-driven cultural norms. This flywheel may in turn reinforce generative AI's aesthetics, as well as the biases these models exhibit. 

Another key aspect of the aesthetics of AI-generated artifacts is the very knowledge that the artifact was created by generative AI, and how that knowledge influences the viewer's perception \cite{moruzzi2020should}. As mentioned above, viewers often engage in “intention guessing,” and the presence of human intention leads to enhanced perceptions of creativity and creative value \cite{do2000intentions, chamberlain2018putting, kruger2004effort, da2015people}. However, as viewers increasingly anthropomorphize generative-AI systems by ascribing them with intention and agency, the credit ascribed to various human actors may change. For instance, we may witness decreasing perceived credit for the human artist or increasing perceived credit for the creator of the technology \cite{epstein2020gets}. Future work should explore ways to quantify and increase output diversity, and study how generative-AI tools may influence aesthetics and aesthetic diversity. In addition, we need new ways of communicating about artist intention in AI production. 

The proliferation of AI-generated content is embedded in a social media landscape where users post content to platforms and these platforms serve content to other users through the filter of opaque, engagement-maximizing recommendation algorithms that leverage personalized patterns gleaned from browsing behavior. The distinct logic of this technological context can shift practices of both production and consumption. To increase visibility on these platforms, creators might continue to prioritize the production of content that satisfies their perceptions of what the algorithms will surface \cite{giblin2022chokepoint, bishop2020algorithmic, bishop_creep}. As both curation algorithms and content creators try to maximize engagement, this may result in further homogenization of content \cite{manovich2018ai}. However, some preliminary experiments \cite{epstein2021social} suggest that incorporating engagement metrics when curating AI-generated content can, in some cases, diversify content. It remains an open question what styles are amplified by recommender algorithms, and how that prioritization affects the types of content creators make and share. Future work must explore the complex, dynamic systems formed by the interplay between generative models, recommender algorithms, and social media platforms, and their resulting impact on aesthetics and conceptual diversity.

\section*{Legal Dimensions of Authorship}
Generative AI’s reliance on training data to automate aspects of creation raises legal and ethical challenges regarding authorship and thus should prompt technical research into the nature of these systems \cite{eshraghian2020human, henderson2023foundation}. Copyright law must balance the benefits to creators, users of generative AI tools, and society at large. In this section, we focus on two distinct (but related) legal challenges. The first is the legal treatment of a model's training data itself, and the second is the legal treatment of the model's  outputs. 

\subsection*{The legal treatment of training data}
AI's use of training data could violate copyright law even before a new output is generated. Four broad proposals have arisen regarding how copyright law could balance the benefits to creators, users of generative-AI tools, and society at large. First, under a highly permissive view, the use of training data could be treated as non-infringing because protected works are not directly copied. Second, the use of training data could be covered by a fair-use exception because a trained AI represents a significant transformation of the training data~\cite{margoni2022deeper, leval1990toward, somepalli2022diffusion, grimmelmann2015copyright, sobel2017artificial,lemley2020fair}.\footnote{It is worth mentioning thought that US fair-use laws are much more permissive than, for example, UK fair-use laws, so ultimately the coverage will depend on jurisdiction.} Third, the use of training data could require an explicit license agreement with each creator whose work appears in the training dataset. A weaker version of this third proposal, is to at least give artists the ability to opt-out of their data being used for generative AI~\cite{huang2023generative}. Finally, a new statutory compulsory licensing scheme that allows artworks to be used as training data but requires the artist to be remunerated could be introduced to compensate artists and create continued incentives for human creation~\cite{licencescomment2020}. The use of training data raises a number of important questions about generative AI:
\begin{enumerate}
\item Does collecting third-party data for training violate copyright? 
\item How often do these models directly copy elements from the training data, versus creating entirely new works \cite{carlini2023extracting, somepalli2022diffusion}?  
\item Even when models do not directly copy from existing works, how should artists' individual styles be protected \cite{shan2023glaze}? 
\item What mechanisms could protect and compensate the artists whose work is used for training these models, or even allow them to opt out, while simultaneously allowing for new cultural contributions from generative AI models? 
\end{enumerate}
Answering these questions and determining how copyright law should treat training data requires substantial technical research to develop and understand the AI systems, social science research to understand perceptions of similarity, and legal research to apply existing precedents to novel technology. Of course, these points represent only an American legal perspective.

\subsection*{The legal treatment of model outputs}

A related but distinct legal question is: who can claim legal ownership over the output of generative AI systems? Answering this requires understanding the creative contributions of a system’s users versus other stakeholders, such as the system’s developers and creators of the training data. AI developers could, for example, claim ownership over outputs through terms of use. In contrast, if a user of a system (e.g., the prompter for text-to-image models or LLMs) has engaged in a meaningfully creative way (e.g., the process is not fully automated, or does not emulate specific works), then they might be considered as the default copyright holders. But how substantial must users’ creative influence be for them to claim ownership? An important exception arises when a major artistic element from the training data or a prompt is part of an output, in which case the artist that owns the relevant work may claim that the output represents a derivative work. How likely is it that major elements from the training data unintentionally appear in the output? Ultimately, answering these questions involve studying not just the models themselves but also the creative process of using AI-based tools. And the answers of these questions may change as users gain more direct control through e.g., painting interfaces. 

Generative AI can also be used to deliberately emulate a specific existing work, either through the use of prompt material or by fine-tuning the AI~\cite{somepalli2022diffusion, baio2022invasive}. The resulting outputs could be characterized as derivative works over which the original artists can claim ownership, although it may also be possible to reward prompt artists through compulsory licenses~\cite{licencescomment2020} or joint ownership~
\cite{jointcomment2020}. Copyright law does not usually protect artistic styles, but artists may legally object to their names being associated with a certain style under misappropriation laws~\cite{baio2022invasive}. Apportioning ownership over outputs requires studying the creative process of using AI-based tools, and may become complex as algorithms provide more direct control to users, for example, through painting interfaces. 

\section{The Labor Economics of Creative Work}
Regardless of legal outcomes, generative AI is likely to transform creative work and employment. Economists have leveraged the skill-biased technological change (SBTC) framework \cite{berman1998implications} to identify which workers may lose employment or wages because of automation \cite{smith2017automation, leontief1952machines, keynes2010economic}. According to this theory, physical routine (repetitive, following an easily-described process, low-skill) workers suffer risks of replacement by technology \cite{acemoglu2011skills, autor2001vthe}. Similarly, the theory also assumes that cognitive non-routine (high-skill) workers are made more productive by technology \cite{acemoglu2011skills, autor2001vthe} and therefore that cognitive and creative workers face less labor disruption from automation because creativity is not readily encodable into concrete rules (i.e., Polanyi's paradox) \cite{frey2017future}. Yet, the new tools have sparked employment concerns for creative occupations such as composers, designers, and writers. This conflict arises because SBTC fails to differentiate between cognitive activities like analytical work and creative ideation. Recent research \cite{frank2019toward, brynjolfsson2018can} demonstrates the need to quantify the specific activities of various artistic workers before comparing them to the actual capabilities of technology. A new framework is needed to characterize the specific steps of the creative process, precisely which and how those steps might be impacted by generative AI tools, and the resulting effects on workplace requirements and activities of varying cognitive occupations \cite{frank2019toward}. For example, human-in-the-loop interactive paradigms could both advance workers' productivity with AI while highlighting areas for future tools to better complement workers~\cite{zhu2018explainable,sanders2008co}.

Although these tools may threaten some occupations, they could increase the productivity of others and perhaps create new ones. For example, historically, music automation technologies enabled more musicians to create—even as earnings skewed \cite{knees2022bias}. Generative-AI systems can create hundreds of outputs per minute which may greatly accelerate the creative process through rapid ideation \cite{leach2022architecture, smith2023trash}. This may reduce production time, and thus reduce costs. In turn, demand for creative work may increase (e.g., the same marketing budget now buys more ads). On the other hand, this acceleration of ideation may undermine aspects of creativity by removing the initial period of prototyping and envisioning associated with a tabula rasa. The impacts on creativity of generative AI tools for ideation requires continued thought and research, yet in either case, the production of creative goods may become more efficient, leading to the same amount of output with fewer workers. Furthermore, some work-for-hire occupations using conventional tools, like illustration or stock photography, might face some displacement. 

Several historical examples bear this out. Most notably, the Industrial Revolution enabled mass scale production of traditionally artisanal crafts (e.g., ceramics, textiles, and steelmaking) with the labor of non-artisans \cite{frey2019technology}. In turn, hand-made goods became treated as luxury items with increased artistic intention \cite{herman3, fuchs2015handmade}. Similarly, when photography reached the mainstream, photographers replaced portraitists and documentary painters. And, the digitization of music removed constraints of learning to physically manipulate instruments and enabled more complex arrangements with many more contributors. By both lowering the barrier to entry while simultaneously making creative tasks more efficient, these tools may change who can work as an artist, in which case employment may rise for artists even as average wages fall.
\section{Impacts on the Media Ecosystem}

As these tools affect creative labor, they also introduce potential downstream harms to the broader information and media ecosystem. As the cost and time to produce synthetic media at scale decreases, the media ecosystem may become particularly vulnerable to impersonation, manipulation, disinformation, and distraction \cite{goldstein2023generative}. New possibilities for the generation of photorealistic synthetic media, for example, may undermine trust in authentically-captured media via the liar’s dividend \cite{chesney2019deep}, and also increase threats of fraud and nonconsensual sexual imagery.  

This begs the question of how to build new methods for building and calibrating trust in media and whether these methods will be used and trusted across contexts and ideologies. Computational tools can provide content provenance and authentication to help preserve information integrity. 

Digital signatures offer an opportunity to actively actively tracking provenance and authentication. For example, visual watermarks could indicate an image's source, but they are easily cropped out or manipulated. The C2PA protocol~\cite{rosenthol2022c2pa} involves cryptographically binding media with provenance metadata on when and where media is recorded and by whom. Cryptographic signatures attached to images in commercial cameras (e.g. Sony) currently serve to authenticate the images the camera captures at the moment of capture. However, digital signatures in metadata is not a panacea because it requires widespread adoption across the media ecosystem. 

A second, complementary, approach relies on post-hoc machine learning and forensic analysis to passively identify statistical and physical artifacts left behind by media manipulation. For example, learning-based forensic analysis techniques use machine learning to automatically detect manipulated visual and auditory content (see e.g. \cite{nguyen2019capsule}). However, these learning-based approaches have been shown to be vulnerable to adversarial attacks \cite{hussain2021adversarial} and context shift \cite{groh2022identifying}. Artifact-based techniques exploit low-level pixel artifacts introduced during synthesis. But these techniques are vulnerable to counter-measures like recompression or additive noise. Other approaches involve biometric features of an individual (e.g., the unique motion produced by the ears in synchrony with speech \cite{agarwal2021detecting}) or behavioral mannerisms \cite{bohavcek2022protecting}). Biometric and behavioral approaches are robust to compression changes and do not rely on assumptions about the moment of media capture, but they do not scale well. However, they may be vulnerable to future generative-AI systems that may adapt and synthesize individual biometric signals.

On social media platforms, people will need to consider whether the media they are consuming may be produced by generative AI. In order to improve people's ability to discern which news headlines are true or false, a body of scholarship has explored the science of misinformation \cite{pennycook2021psychology}. This work has focused primarily on why and how people come to believe misinformation and why they share it. Pennycook and Rand address these questions by examining the role of reasoning and heuristics in people's ability to discern truth from falsehood \cite{pennycook2021psychology}. Fundamentally, people can accurately distinguish between real and fake news when they are paying attention to accuracy, but distraction by social motivations can undermine discernment \cite{pennycook2020understanding,epstein2023social}. Indeed, people's attention is often focused away from accuracy when browsing social media and instead attention is often oriented towards moral \cite{brady2020mad}, emotional \cite{brady2017emotion, bergermilkman} and sensational content \cite{epstein2022quantifying}. As a result, misinformation may spread faster than accurate information \cite{vosoughi2018spread} (though see caveats \cite{juul2021comparing}). This body of scholarship raises the question: how will synthetic media impact the extent to which people are susceptible to and spread misinformation? 	

Preliminary evidence suggests that photographs can influence people's susceptibility to fake news headlines without providing any direct visual evidence of the headlines' claims \cite{newman2012nonprobative}. While visual information is more powerful than textual information for inciting believability in content, this only translates to a small increase in persuasiveness \cite{wittenberg2021minimal}. Furthermore, recent research reveals that the visual components of synthetic media can be useful for identifying its synthetic origins \cite{groh2022deepfake}.

There are many other potential impacts of generative AI on the information environment beyond the potential for explicitly faked photorealistic imagery or plausible-sounding audio. Large language models (LLMs) that assist creators can generate fluent written content, without reliable information verification in the output, or in service of a particular ideology \cite{jakesch}. While the focus of this paper--artistic creation--bleeds into these broader trends in content creation writ large, these capabilities of LLMs ultimately introduce a set of issues distinct from image generation that are beyond the scope of this paper. While the approaches to provenance and watermarking technologies discussed above may be useful for mitigation, future work is needed to properly diagnose the impact of LLMs on the information environment and propose solutions. 

With the proliferation of both AI-generated visual and written media, another principal concern is how the increased amount of information will impact online environments. Lorenz at al. \cite{lorenz2019accelerating} find that increases in the amount of information available can decrease collective attention spans. The explosion of AI-generated content may in turn hamper society’s ability to engage in collective discussion and action in important arenas such as climate and democracy. Furthermore, generative AI systems may allow authoritarian governments or bad actors to mass produce articles, blogs, and memes that drown out organic public discourse.

These concerns and approaches raise important research questions:
\begin{enumerate}
    \item What is the role of platform interventions such as tracking source provenance and detecting synthetic media downstream for governance and promoting trust \cite{farid2022creating}? 
    \item How does the proliferation of synthetic media affect trust in real media, such as unedited journalistic photographs? 
\end{enumerate}
 \begin{figure}
    \centering
    \includegraphics[width=0.99\textwidth]{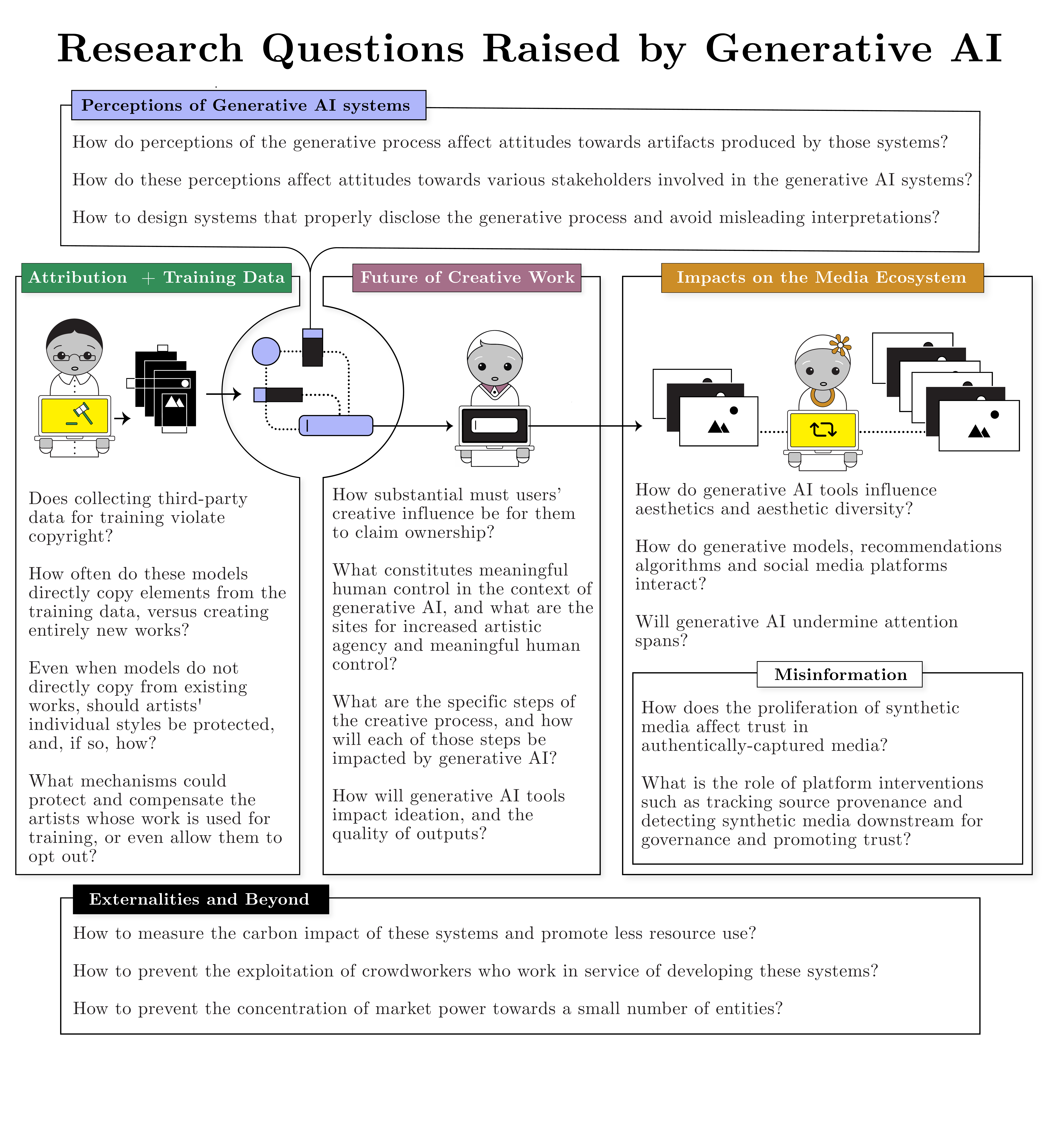}
    \caption{Research questions raised by generative AI}
    \label{fig:ngrams}
\end{figure}

\section*{Discussion and Limitations}

There are other potential societal impacts of generative AI that we did not cover, particularly those related to the externalities of using these models (see \cite{epicharms} for a review). For one, there is the potential of such models to exacerbate anthropogenic climate change  \cite{strubell2019energy, epicharms}, with these environmental impacts disproportionately affecting marginalized populations\cite{westra2001faces, bender2021dangers}. This requires new ways to measure the carbon impact of these systems \cite{lacoste2019quantifying, anthony2020carbontracker,henderson2020towards} and increased scrutiny toward the energy usage of these systems \cite{bender2021dangers, epicharms}. In addition, it is important to consider the exploitation of crowd workers who work in service of developing these systems by labeling data: their conditions include low pay, dehumanizing tasks, and workplace surveillance \cite{gray2019ghost, kalluri2021don}. In addition, these models may concentrate market power towards a small number of entities \cite{zuboff2015big}. There are but a few of the additional harms that generative AI may induce, and comprehensively mapping those is important future work \cite{epicharms}.

We hope that generative AI will incite more research into the nature of human creativity. Just as machine learning research has led to advances in neuroscience and cognitive science \cite{savage2019ai}, the development of generative-AI tools could help us understand more about the nuances of human creativity. By definition, these generative models are trained by reducing error on reconstruction tasks on training data. As such, they are fundamentally bound by reproducing what they have already seen. Therefore, a key future direction is novel algorithmic methods for embedding improvisation in AI systems \cite{broad2021active, zammit2022seeding,hertzmannICCC}. This can also be achieved by using generative AI earlier in the workflow for speculation and idea generation \cite{epstein2022happy, colton2021generative, smith2023trash}, or by building algorithms that are explicitly designed to interact with distinct modes of human creativity.

Every artistic medium mirrors and comments on the issues of its time, and contemporary AI-generated art  reflects present issues surrounding automation, corporate control, and the attention economy. Ultimately, we express our humanity through art, so understanding and shaping the impact of AI on creative expression is at the center of broader questions about its impact on society. 

The widespread adoption of generative AI is not inevitable. Rather, its uses and impacts will be shaped by the collective decisions made by technology developers, users, regulators and civil society. Therefore, new research into generative AI is required to ensure the use of these technologies is beneficial and must engage with critical stakeholders, particularly artists and creative laborers themselves, many of whom actively engage with difficult questions at the vanguard of societal change. 

\section*{Contributions and Position}
This paper is a large-scale collaboration between the $14$ authors. It was initiated and conceptualized by Ziv Epstein, Aaron Hertzmann, Amy Smith and Hope Schroeder at the 2023 International Conference of Computational Creativity. Co-authors were invited with expertise in each of the four themes: Shifts in Culture \& Aesthetics, Legal Dimensions of Authorship, Labor Economics of Creative Work, and Impacts Media Ecosystem. The Shifts in Culture \& Aesthetics section was led by Laura Herman with supervision from Memo Akten and contribution from Memo Akten, Olga Russakovsky, Neil Leach, and Ziv Epstein. The Legal Dimensions of Authorship section was led by Robert Mahari with supervision from Jessica Fjeld and contribution from Jessica Fjeld and Aaron Hertzmann. The Labor Economics of Creative Work section was led by Morgan R. Frank with contribution from Alex Pentland and Amy Smith. The Impacts on the Media Ecosystem section was led by Matthew Groh and Hope Schroeder with supervision from Hany Farid and contribution from Hany Farid and Ziv Epstein. The drafts were collectively edited and refined by all authors.

This work is an academic collaboration, but we  acknowledge that two authors---Aaron Hertzmann and Laura Herman---work for Adobe, which makes generative AI tools. These authors collaborated on this piece building on their scholarship on these topics, and both have joint academic affiliations. Furthermore, three authors---Ziv Epstein, Matthew Groh and Morgan R. Frank---had previously consulted for OpenAI by red teaming their generative AI systems (Ziv Epstein and Matthew Groh red teamed DALL-E 2 and Morgan R. Frank red teamed chatGPT). 
\end{singlespace}


\bibliographystyle{unsrt}
\begin{singlespace}
\bibliography{scibib}
\end{singlespace}


\end{document}